\documentclass{article}

    \PassOptionsToPackage{numbers, sort, comma, square}{natbib}

\usepackage[final]{neurips_2022}

\usepackage[utf8]{inputenc} 
\usepackage[T1]{fontenc}    
\usepackage{hyperref}       
\usepackage{url}            
\usepackage{booktabs}       
\usepackage{amsfonts}       
\usepackage{nicefrac}       
\usepackage{microtype}      
\usepackage{xcolor}         

\usepackage{graphicx,wrapfig}
\usepackage{enumitem}
\usepackage{amsmath}
\usepackage{amssymb}
\usepackage{subfig}
\usepackage{soul}

\usepackage{tikz}
\usetikzlibrary{fit,positioning}

\usepackage{dsfont}

\usepackage{graphicx} %
\usepackage{amssymb}
\usepackage{times}
\usepackage{epsfig}
\usepackage{graphicx}
\usepackage{color}
\usepackage{bbm}
\usepackage{multicol}
\usepackage{natbib}
\usepackage{wrapfig}

\providecommand{\hide}[1]{}

\usepackage{rotating}

\usepackage{amsthm}
\usepackage{amsopn,amssymb,thmtools,thm-restate}

\usepackage{bm} %
\usepackage{algorithm}%
\usepackage{algpseudocode}%
\newcommand{\vct}[1]{\boldsymbol{#1}} %

\newcommand{\ProbOpr}[1]{\mathbb{#1}}

\newcommand{\expect}[2]{%
\ifthenelse{\equal{#2}{}}{\ProbOpr{E}_{#1}}
{\ifthenelse{\equal{#1}{}}{\ProbOpr{E}\left[#2\right]}{\ProbOpr{E}_{#1}\left[#2\right]}}} %
\newcommand{\var}[2]{%
\ifthenelse{\equal{#2}{}}{\ProbOpr{VAR}_{#1}}
{\ifthenelse{\equal{#1}{}}{\ProbOpr{VAR}\left[#2\right]}{\ProbOpr{VAR}_{#1}\left[#2\right]}}} %

\newcommand{\vx}{{\vct{x}}}

\newcommand{\vb}{\vct{b}}

\newcommand{\vh}{\vct{h}}

\newcount\Comments  %
\newcommand{\kibitz}[2]{\ifnum\Comments=1\textcolor{#1}{#2}\fi}
\usepackage{color}
\definecolor{red}{rgb}{1,0,0}

\newcount\Edits  %
\newcommand{\edit}[1]{\ifnum\Edits=1\textcolor{blue}{#1}\else{#1}\fi}

\title{Multi-step Planning for Automated Hyperparameter Optimization with OptFormer}

\author{%
  Lucio M. Dery\thanks{Work done whilst interning at Deepmind. Correspondence to \texttt{<ldery@andrew.cmu.edu>} or \texttt{<yutianc@deepmind.com>}} \\
  Carnegie Mellon University\\
   \And
   Abram L. Friesen \\
   Deepmind \\
   \AND
  Nando De Freitas \\
  Deepmind \\
   \And
  Marc'Aurelio Ranzato \\
  Deepmind \\
   \And
   Yutian Chen \\
   Deepmind \\
}

\begin{document}

\maketitle

\begin{abstract}
As machine learning permeates more industries and models become more expensive and time consuming to train, the need for efficient automated hyperparameter optimization (HPO) has never been more pressing. Multi-step planning based approaches to hyperparameter optimization promise improved efficiency over myopic alternatives by more effectively balancing out exploration and exploitation. However, the potential of these approaches has not been fully realized due to their technical complexity and computational intensity. In this work, we leverage recent advances in Transformer-based, natural-language-interfaced hyperparameter optimization to circumvent these barriers. We build on top of the recently proposed OptFormer which casts both hyperparameter suggestion and target function approximation as autoregressive generation thus making planning via rollouts simple and efficient.  We conduct extensive exploration of different strategies for performing multi-step planning on top of the OptFormer model to highlight its potential for use in constructing non-myopic HPO strategies.
\end{abstract}

\section{Introduction}
\setcounter{footnote}{0} 

The performance of a machine learning (ML) model is deeply tied to the choice of hyperparameters used for its training and deployment. As such, the problem of automated hyperparameter optimization (HPO) has become increasingly relevant as ML models have become more ubiquitous \citep{li2020system, chen2022towards, yang2022tensor}. Given a function that describes a model's performance, $f(\vx)$\footnote{Here, performance captures the metric that we wish to optimize, such as accuracy, runtime, etc.}, and a space of hyperparameters $\mathcal{X}$, the goal of HPO is to find $\vx^{*} = \mathrm{argmax}_{\{\vx \in \mathcal{X}\}} f(\vx)$ in finite time ($T$) by making a sequence of decisions $\mathcal{S} = \{\vx_{1}, \ldots, \vx_{T}\}$ such that $\vx^{*} \in \mathcal{S}$.

Popular approaches to HPO are based on Bayesian optimization (BO) \citep{brochu2010tutorial, shahriari2015taking}. In BO, one builds a time-dependent surrogate model $\tilde{f}_{t}(\vx)$ using evidence collected from past evaluations of the true function $\{y_1 = f(\vx_1), \ldots, y_{t - 1} = f(\vx_{t - 1})\}$.  Most methods proposed under this framework proceed as follows: at any timestep $t$, 
they suggest $\vx_{t} \approx \mathrm{argmax}_{\vx \in \mathcal{X}} \alpha(\vx; \tilde{f}_t)$ where $\alpha$ is called an acquisition function that measures the expected utility after one additional evaluation of $f$ at $\vx$ \citep{shahriari2015taking, wu2016parallel, wang2017max}. These methods are myopic. They make decisions exclusively based on the estimated function at the current timestep $\tilde{f}_t$ and can therefore be sub-optimal since they do not account for the impact of the current decision on future evaluations. Planning-based HPO attempts to avoid this by appropriately balancing exploration and exploitation during decision making. They attempt to maximize the long term reward over a time horizon, $h$, by considering the influence of the choice of $\vx_{t:t+h-1}$ on $\tilde{f}_{t + h}$ through multistep lookahead \citep{gonzalez2016glasses, lam2017lookahead, wu2019practical}. Such methods promise improved query efficiency -- needing fewer configuration evaluations to achieve threshold performance -- and improved performance of the model after all evaluations are considered when compared to myopic approaches. As recent ML models like pre-trained language and vision models become larger and thus both more expensive and time consuming to train \citep{brown2020language, hoffmann2022training, chowdhery2022palm}, the above promises have become more enticing. 

Despite these promises, planning-based HPO has not been widely adopted. This is because suggested approaches have been both technically complex to implement and computationally intensive to execute. In this paper, we circumvent both these challenges by leveraging recent advances in Transformer based \citep{vaswani2017attention}, natural language interfaced HPO. Specifically, \citet{chen2022towards} recently proposed an approach dubbed OptFormer which presents a unified, natural language interface for hyperparameter optimization and thus allows transfer learning between HPO problems. Relevant to us is the structure of the OptFormer model: it jointly performs both hyperparameter suggestion $\vx_t$ and function prediction $\tilde{f}_t(\vx_t)$ as autoregressive generation. 

By leveraging the ease and speed of autoregressive generation with OptFormer, we present a novel planning-based HPO algorithm that is simple and scalable; thus bypassing the typical challenges of traditional methods.  We incorporate multi-step lookahead into both candidate generation (instead of doing a global search as typical of most HPO approaches) and candidate evaluation (we provide a planning-based acquisition function $\alpha$) since performing rollouts with OptFormer is simply multistep generation for $h$ timesteps. In Section \ref{section:method}, we present our method in detail and in Section \ref{section:results_and_disc} we empirically validate our approach under two settings: the BBOB dataset \citep{elhara2019coco}, which is a synthetic benchmark, and the RealWorldData dataset \citep{chen2022towards}, which is constructed from aggregating actual hyperparameter tuning experiments. Our results highlight that incorporating planning into transformer-based hyperparameter optimization models is a promising research direction and we invite the community to explore this setting further.

\section{Related Work}

As already discussed, most approaches to hyperparameter optimization leverage Bayesian optimization (BO). These methods typically build a Gaussian process (GP) model \cite{rasmussen2003gaussian, shahriari2015taking} as the surrogate $\tilde{f}_{t}(\vx)$ and proceed to suggest $\vx_{t} \approx \mathrm{argmax}_{\vx \in \mathcal{X}} \alpha(\vx; \tilde{f}_t)$ where $\alpha$ is an acquisition function.  An acquisition function measures the expected utility after one additional evaluation of $f$ at $\vx$. Past work has explored a variety of acquisition functions such as probability of improvement (PI) \citep{shahriari2015taking}, expected improvement (EI) \citep{mockus1978application, jones1998efficient} and upper confidence bound (UCB) \citep{agrawal1995sample, auer2002using}. These acquisition functions are myopic since they depend solely on the surrogate at the current timestep and so do not directly account for the impact of a greedy choice on downstream evaluations.

Unlike myopic HPO methods, planning based approaches fundamentally require building models of the future to assess the impact of a current decision on later timesteps. Though these methods also rely on a GP as a surrogate model, each point in multi-step planning involves \textit{fantasizing/imagining} an updated GP posterior $\left(\tilde{f}_{t + 1} \mid_{\tilde{\vx}_t}\right), \ldots, \left(\tilde{f}_{t + h} \mid_{\tilde{\vx}_t, \tilde{\vx}_{t + 1}, \ldots, \tilde{\vx}_{t + h - 1}}\right)$ based on simulated choices from lookaheads $\{(\tilde{\vx}_t, \tilde{y}_t), \ldots, (\tilde{\vx}_{t + h - 1}, \tilde{y}_{t + h-1})\}$ \citep{lam2016bayesian, jiang2020efficient}. Note that we use $\tilde{\vx}_{t}$ to represent a fantasized decision, while $\vx_{t}$ is the actual choice made at timestep $t$. Whilst multi-step planning is promising, constructing the posterior of a GP model requires matrix inversion which is a compute-intensive operation \citep{cormen2022introduction}. Even outside of this limitation, traditional planning based approaches are compute intensive due to 
(i) poor scaling behavior of the search tree---$\mathcal{O}(q^{h})$ where $q$ is the number of choices at each decision point for each lookahead step \citep{lam2016bayesian, lam2017lookahead}---which forces most methods to explore short horizons, typically $h \in \{1, 2\}$, and
(ii) nested expectation and maximization: marginalizing future observation $\tilde{y}_{t+j}, j<h$ and global search on the acquisition function to obtain query $\tilde{\vx}_{t+j}$ at every lookahead step. Another issue is complexity of implementation. In an attempt to reduce the computational burdens described above, complex approaches have been suggested that require significant expertise to execute properly. For example \citet{jiang2020efficient} attempt to overcome the problem of nested expectation and maximization by leveraging the reparameterization trick \citep{kingma2013auto} to couple sampling and optimization. They build a differentiable objective over a multi-step tree and then optimize over this objective to get a proposal for the next suggestion $\vx_t$; a far from simple method. Rollout policy \citep{bertsekas2012dynamic} provides a tractable alternative to approximate the nested expectation by playing multiple rollouts using a base policy but maximization in the search space is still required at every step in the base policy.

Recently \citet{chen2022towards} proposed a transformer-based \citep{vaswani2017attention} model called OptFormer, which presents a unified, natural language interface for HPO and thus allows transfer learning between HPO problems. OptFormer jointly performs both hyperparameter suggestion $\tilde{\vx}_t$ and function prediction $\tilde{f}_t(\tilde{\vx}_t)$ as autoregressive generation. This means that the OptFormer model does not suffer from significant computational expense during updates in response to fantasized decisions. Also, since the OptFormer model has been trained to mimic base policies like GP-UCB, autoregressive hyperparameter generation serves as a fast and strong approximation of the inner maximization step discussed in the preceding paragraph. The model is also simple to use and, due to the widespread popularity of transformer models \citep{lin2021survey, khan2021transformers}, is accessible to a wide audience of ranging technical expertise. The absence of the computational and technical constraints typical of non-myopic HPO means the OptFormer model is uniquely positioned as a fertile test-bed for exploring practical planning based hyperparameter optimization. 

\section{Background on OptFormer}
\label{section:optformer}

OptFormer is a transformer-based hyperparameter optimization method proposed in \citet{chen2022towards}. It converts historical hyperparameter tuning experiments, each consisting of a task description, $m$, and optimization trajectory, $\vh_t = ((\vx_1, y_1), \dots, (\vx_t, y_t))$, into a large dataset of over 700K text sequences, and then trains a transformer to learn (1) $P(\vh_t|m)$, the prior distribution of the data generative process; (2) $P(\vx_t|m, \vh_{t-1})$, the prior conditional distribution, which corresponds to the distribution of the next query made by the base HPO algorithms (eg. GP-UCB) that generated the tuning trajectory data (equivalently, the behavior policy in reinforcement learning); and (3) $P(y_t|m, \vh_{t-1}, \vx_t)$, the predicted distribution of the unknown function value $f$ at input $\vx_t$ given observations $\{(\vx_1, y_1), \dots, (\vx_{t-1}, y_{t-1})\}$---the surrogate model $\tilde{f}_t(\vx)$ required by traditional BO methods. Note that all OptFormer distributions are modelled autoregressively. Thus, generating $\vx_{t}$ is achieved by simply prompting OptFormer with a text trajectory $(m; \vh_{t-1})$ to produce the distribution $P(\vx_t|m, \vh_{t-1})$ at the model's output and then sampling from it. Once $\vx_{t}$ is generated, it is appended to the trajectory $(m; \vh_{t-1}; \vx_{t})$ which we autoregressively feed back into OptFormer to obtain $P(y_t|m, \vh_{t-1}, \vx_t)$.

\begin{figure}[h!]
    \vspace{-1mm}
    \centering
    \includegraphics[scale=0.15]{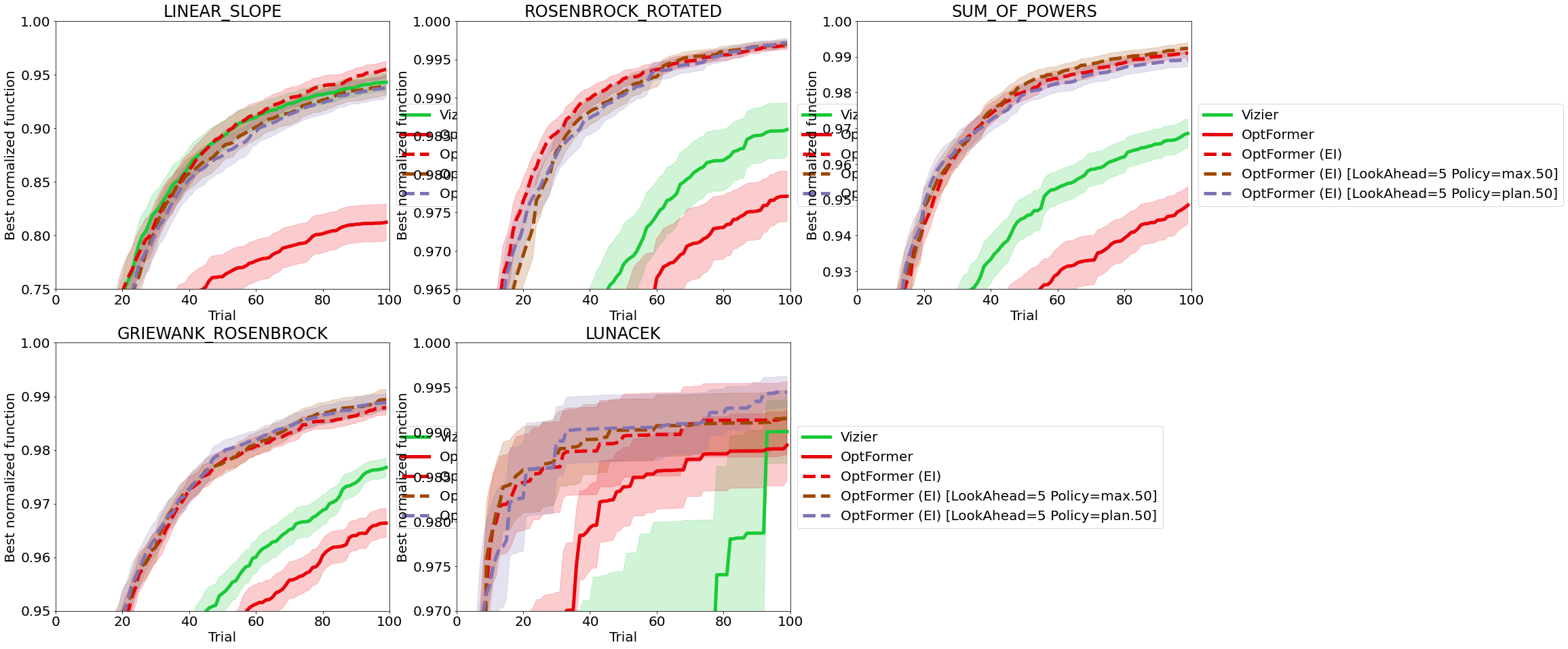}
    \caption{Performance on 5 test functions from the BBOB dataset. OptFormer is the unaugmented prior policy (we just sample $\vx_1 = P(\tilde{\vx}_{t} | m, \vh_{t - 1})$ which we play).}
    \label{fig:bbob_results}%
    \vspace{-2mm}
\end{figure}

OptFormer proposes two policies for hyperparameter optimization. First, a \textbf{prior policy} that samples hyperparameters $\vx_t$ autoregressively from the prior conditional distribution directly. Second, a two-stage \textbf{augmented policy} as follows:
\begin{enumerate}
    \item{ \textbf{Candidate generation}: proposes a set of $n$ hyperparameter candidates from the prior policy. The OptFormer model performs autoregressive generation conditioned on the evidence and problem meta-data $m$, producing a set of $n$ possible candidates $C_{t} = \{ \tilde{\vx}^{(i)}_{t} ~|~ \tilde{\vx}^{(i)}_{t} \sim P(\vx_{t}|m, \vh_{t - 1}) \}_{i \in [n]}$. This step is absent from traditional BO approaches. Since OptFormer has been trained to mimic prior policies, $C_{t}$ is typically a high quality set of suggestions.
    }
    \item{\textbf{Candidate evaluation}: computes a 1-step acquisition function $\alpha(\vx_t^{(i)};\tilde{f}_t(\vx))$ such as UCB or expected improvement (EI) based on its own surrogate model $\tilde{f}_t(\vx) = P(y_t|m, \vh_{t-1}, \vx_t)$. OptFormer suggests the candidate $\vx_{t} = \mathrm{argmax}_{\vx \in C_{t}} \alpha(\vx, \tilde{f}_t(\vx)) $.
    }
\end{enumerate}
Whilst the prior policy just mimics the base HPO algorithm, the augmented policy constitutes a 1-step policy improvement over the base algorithm and empirically yields better performance compared to the prior policy \citep{chen2022towards} (Figure \ref{fig:bbob_results}).

The OptFormer procedure described above uses a myopic acquisition function and does not leverage simulations to gauge the effects of the current choice on future evaluations. The improvements achieved by the OptFormer augmented policy over its base prior engender hope that further performance gains can be realised by incorporating planning.
\section{Planning-based HPO with OptFormer}
\label{section:method}
In this work, we develop approaches that are able to plan over a given horizon $h$. Below, we describe the modifications we make to the two-stage OptFormer process to perform multi-step planning. For notational clarity, we use $\tilde{\vb}_t$ to represent a fantasized/imagined quantity at timestep $t$ and $\vb_{t}$ for the actual observed or decided quantity.

\begin{figure}[t!]
    \centering
    \vspace{-4mm}
    \includegraphics[scale=0.16]{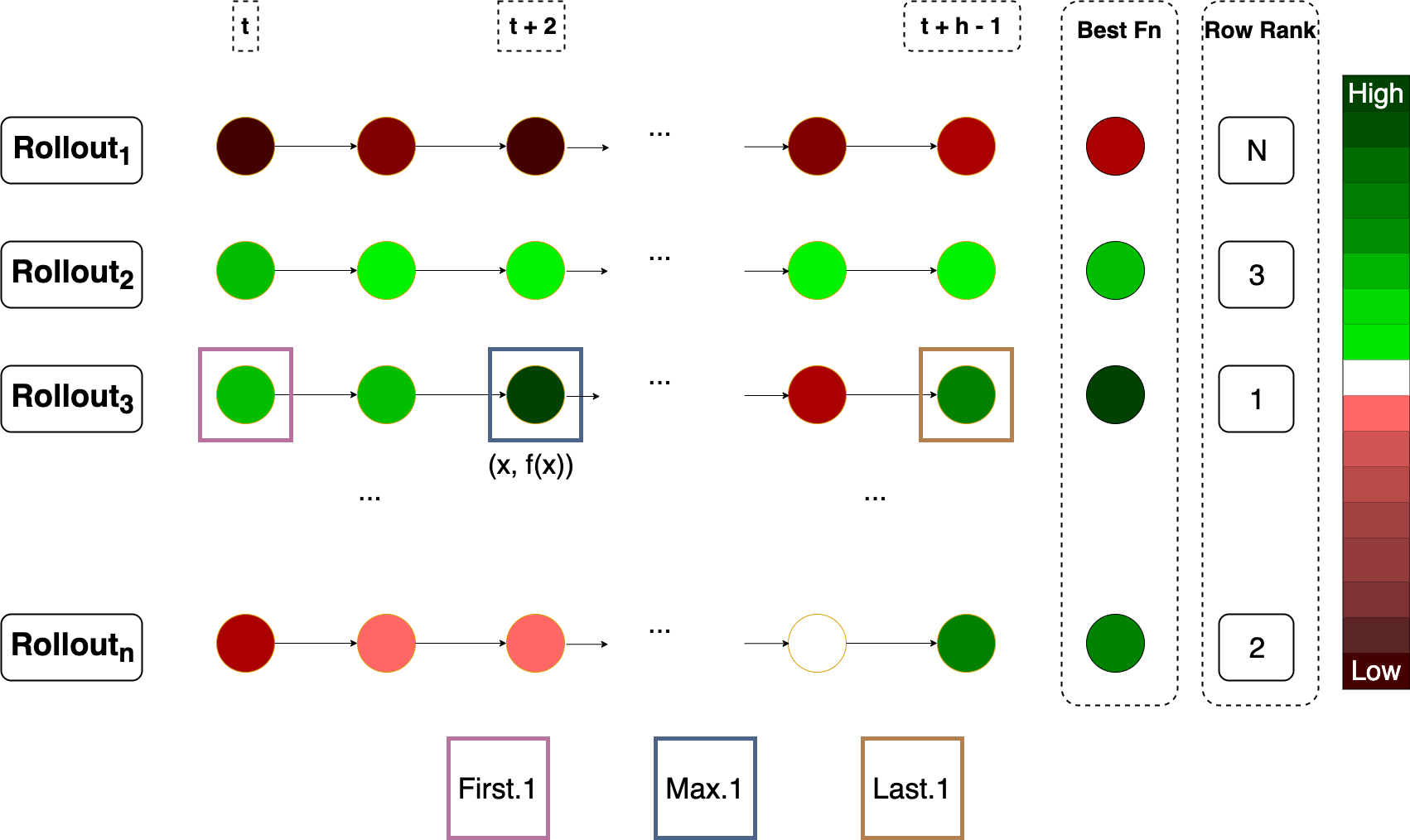}
    \caption{We refine $C_{t}$ by performing roll outs with each candidate. We rank trajectories according to the highest sampled $\tilde{y}$ within the trajectory -- this gives us the trajectory's row rank. Given the trajectories generated, we propose 3 criteria for refining $C_{t}$.  (1) First.k picks k of the seed hyperparameters in $C_t$ that have the highest row rank. (2) Max.k picks the top k $\tilde{\vx}$ that led to the highest $\tilde{y}$ within the top k row ranked trajectories. (3) Last.k picks the top k members of column $t + h - 1$, sorted according to row rank }%
    \label{fig:cand_gen}%
\end{figure}

\subsection{Candidate Generation}
\label{subsect:cand_gen}
As mentioned above, the set $C_{t}$ from the OptFormer prior policy is a strong baseline of potential suggestions. However, we can further leverage our ability to autoregressively prompt the OptFormer model to generate a new horizon-aware set $C^{h}_{t}$, that refines $C_{t}$ by fantasizing future time-steps. Specifically, given $\tilde{\vx}_{t} \in C_{t}$, we can roll out the prior policy to generate \textit{a single} length $h$ sequence of fantasized hyperparameters and function values $\tilde{\vh}(\tilde{\vx}_t) = \{(\tilde{\vx}_t, \tilde{y}_{t}), \ldots, (\tilde{\vx}_{t + h-1}, \tilde{y}_{t + h-1})\}$ by repeatedly sampling from the OptFormer and then autoregressively prompting it with the sampled values: 
\begin{equation*}
    \begin{split}
        \tilde{\vx}_{t + j} &\sim P(\vx_{t + j} ~|~m, \vh_{t - 1}, \tilde{\vx}_{t}, \ldots, \tilde{y}_{t + j - 1}) \quad \text{where } j \in [1, \ldots h] \\
        \tilde{y}_{t + j} &\sim P(y_{t + j} ~|~m, \vh_{t-1}, \tilde{\vx}_{t}, \ldots, \tilde{y}_{t + j - 1}, \tilde{\vx}_{t + j}) \quad  \text{where } j \in [1, \ldots h]
    \end{split}
\end{equation*}
Given these sequences (one for each member of $C_{t}$), we can select $k \leq n$ rollout-aware hyperparameter candidates. Consider a rollout trajectory $\tilde{\vh}(\tilde{\vx}^{i}_t)$ seeded by some $\tilde{\vx}^{i}_t \in C_{t}$. Let $\mathcal{Y}(\tilde{\vh})$ be the set of sampled function values $\tilde{y}$ that appear in the trajectory and $\mathcal{X}(\tilde{\vh})$ be the corresponding set of sampled hyperparameter suggestions. We can rank each rollout based on the highest fantasized function value $\tilde{y}^{\ast}(\tilde{\vx}^{i}_t) =  \max\left(\mathcal{Y}(\tilde{\vh}(\tilde{\vx}^{i}_t))\right)$ to obtain the row rank of the rollout (See Figure \ref{fig:cand_gen}). We can then define the following set of rollout trajectories based on which we will extract $C^{h}_{t}$:
$$ \tilde{\mathbf{H}}(k) = \mathrm{argtopk}_{\{\tilde{\vh}(\tilde{\vx}^{i}_t)\}} ~~ \tilde{y}^{\ast}(\tilde{\vx}^{i}_t) $$
Interpreted in terms of Figure \ref{fig:cand_gen}, this corresponds to taking the top $k$ trajectories as defined by row rank (and we define the row-rank by comparing the best fantasized $\tilde{y}$ achieved by each rollout). Having constructed $\tilde{\mathbf{H}}$, we suggest three different criteria as follows: \\
1. \textbf{First.k: } $C^{h, F}_{t} = \{\tilde{\vx}^{i}_{t} \in C_t ~|~ \tilde{\vh}(\tilde{\vx}^{i}_{t}) \in \tilde{\mathbf{H}} \}$.\\
Intuitively, this approach selects the top $k$ entries \textit{in the original set $C_{t}$} that led to the highest fantasized function value $\tilde{y}$ during rollouts. We posit that these candidates may be the most promising to explore further within $C_t$. Figure \ref{fig:cand_gen} is a pictorial rendering of the approach.

2. \textbf{Max.k: } $C^{h, M}_{t} = \{\tilde{\vx}^{\ast}(\tilde{\vx}_t^{i}) \in \mathcal{X}(\tilde{\vh}_t(\tilde{\vx}_t^{i}) ~|~ \tilde{\vh}(\tilde{\vx}^{i}_{t}) \in \tilde{\mathbf{H}}\}$ \\ 
Here $\tilde{\vx}^{\ast}(\tilde{\vx}_t^{i})$ is the hyperparameter suggestion that directly led to $\tilde{y}^{\ast}(\tilde{\vh})$ within its trajectory.
The Max.k approach selects the hyperparameter suggestion that led to the highest fantasized function value $\tilde{y}$ within each of the highest ranked trajectories $\tilde{\mathbf{H}}$ (See Figure \ref{fig:cand_gen}).

3. \textbf{Last.k: } $C^{h, L}_{t} = \{\tilde{\vx}_{t + h - 1}(\tilde{\vx}_t^{i}) \in \mathcal{X}(\tilde{\vh}_t(\tilde{\vx}_t^{i}) ~|~ \tilde{\vh}(\tilde{\vx}^{i}_{t}) \in \tilde{\mathbf{H}}\}$ \\
$C^{h, L}_{t}$ contains the last sampled $\tilde{\vx}_{t + h - 1}$ during rollouts within the highest ranked trajectories $\tilde{\mathbf{H}}$. This criterion is motivated by the fact that OptFormer was trained to mimic a base prior policy. Since the prior policy used to construct OptFormer training data improves over time (due to access to more observations), we posit that hyperparameter suggestions made by OptFormer from later rollout timesteps are more promising candidates.
\begin{figure}[h!]
    \centering
    \includegraphics[scale=0.19]{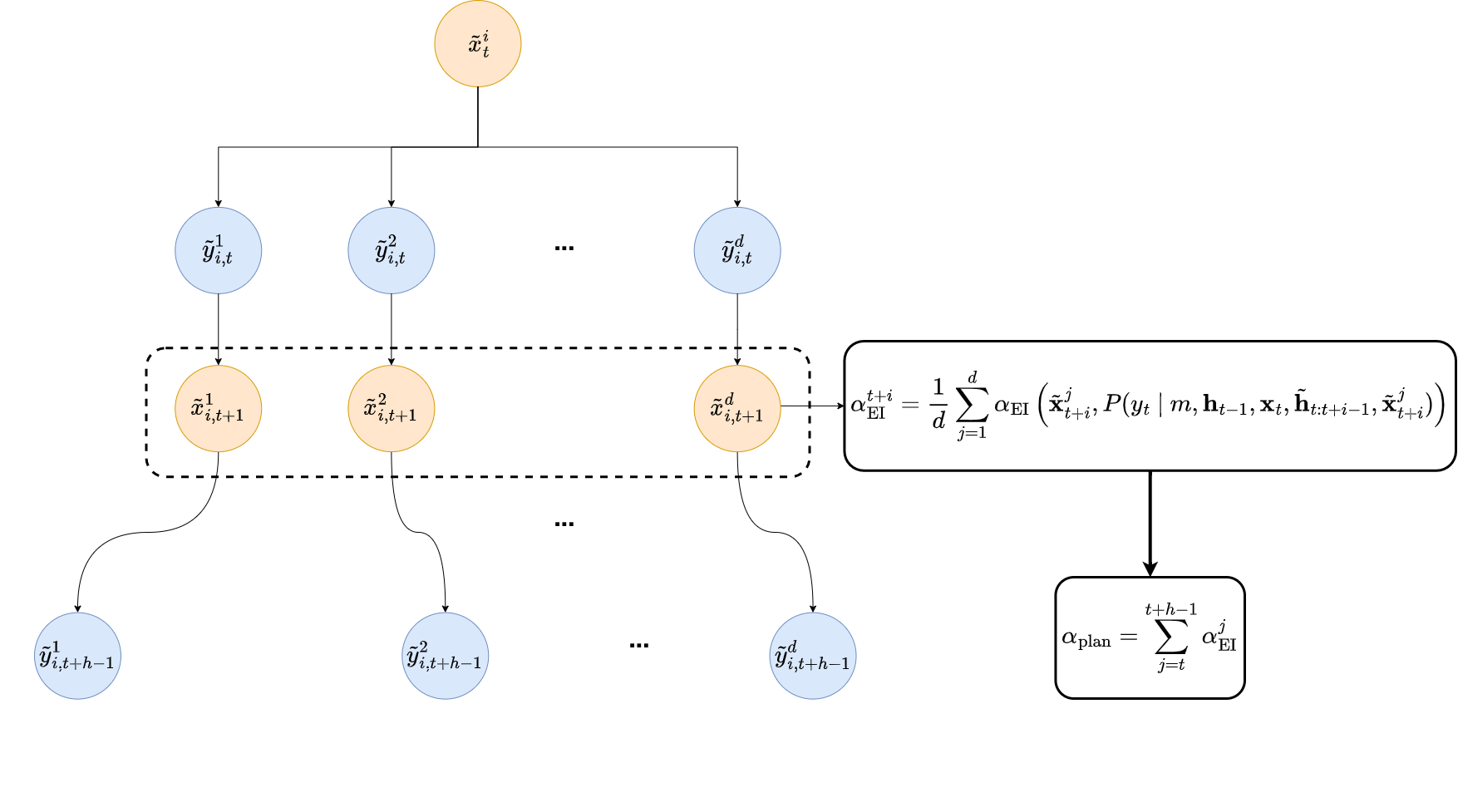}
    \vspace{-3mm}
    \caption{An example instantiation of our planning-based acquistion function. For each candidate $\tilde{\vx}^{i}_t \in C^{h}_{t}$, we perform $d$ roll outs of length $h$ each. At each timstep during rollouts, we compute the expected improvement (EI) and average over all trajectories. The final acquisition function is the sum of the averaged EI over the full trajectory horizon $h$.}%
    \label{fig:plan}%
    \vspace{-5mm}
\end{figure}
\subsection{Candidate Evaluation}
\label{subsect:cand_eval}
Having constructed a refined set $C^{h}_{t}$, it remains to decide which member of the set to suggest. Here, we investigate two acquisition functions: 
\begin{enumerate}[leftmargin=*]
    \item Expected improvement (EI): We suggest $\vx = \mathrm{argmax}_{\tilde{\vx}_t \in C^{h}_{t}} \ \alpha_{\mathrm{EI}}(\tilde{\vx}_t, P(y_{t} ~|~ m, \vh_{t-1}, \tilde{\vx}_{t}))$ where $\alpha_{\mathrm{EI}} = \mathbb{E}_{\tilde{y} \sim P(y_{t} ~|~ m, \vh_{t-1}, \tilde{\vx}_{t})} \left[\max(\tilde{y} - y^{*}_{t - 1}, 0)\right]$ is the expected improvement acquisition function and $y^{*}_{t - 1} = \max\left(\mathcal{Y}(\vh_{t - 1})\right)$ is the highest observed return so far. $\mathcal{Y}(\vh_{t - 1})$ is as defined in Section \ref{subsect:cand_gen}. We can easily compute the expectation above given that the transformer outputs a distribution over possible values of $\tilde{y}_{t}$. Note that EI on its own is myopic, we use $C^{h, *}_{t}$ as our conduit for incorporating planning information.
    \item Planning based EI: For each candidate $\tilde{\vx}_t$ we perform $d$ rollouts each for $h$ timesteps. At each timestep, we can compute $\alpha_{\mathrm{EI}}$ based on the simulated trajectory so far and average across rollouts.
    $$\alpha_{\mathrm{Plan}}(\tilde{\vx}_t) =  \sum^{h - 1}_{i = 0} \frac{1}{d} \sum_{j}^{d} \alpha_{\mathrm{EI}}\left(\tilde{\vx}^{j}_{t + i}, P(y_{t} ~|~ m, \vh_{t - 1}, \tilde{\vx}_t, \tilde{\vh}^{i}_{t}, \tilde{\vx}^{j}_{t + i}) \right)$$
    We then suggest $\vx = \mathrm{argmax}_{\tilde{\vx}_t \in C^{h}_{t}} \ \alpha_{\mathrm{Plan}}(\tilde{\vx}_t)$ as the next hyperparameter to evaluate. Figure \ref{fig:plan} demonstrates the details of computing $\alpha_{\mathrm{Plan}}$. By performing $d$ rollouts, we can average $\alpha_{\mathrm{EI}}$ at every rollout timestep and thus get a lower variance estimate. Also note that whilst $\alpha_{\mathrm{Plan}}$ generates $\mathcal{O}(dh)$ trajectories per candidate, classic tree based multistep planning HPO methods would require $\mathcal{O}(d^{h})$.
\end{enumerate}

In this section, we have presented augmentations to the OptFormer model that enable it to perform multistep planning. Next, we will setup an experimental framework within which we will seek to validate our proposed changes.

\section{Experimental Setting}

\textbf{Datasets}: OptFormer leverages past HPO trajectories for supervised training of the transformer-based model. \citet{chen2022towards} introduce the  RealWorldData dataset, based on the Google Vizier \citep{golovin2017google} database, which they use to train OptFormer. The RealWorldData has 750K trajectories---each with 300 trials on average ---and covers many HPO tuning experiments run at Google across domains like vision, speech, NLP, and robotics. We will use this dataset along with the blackbox optimization benchmark (BBOB) dataset
\citep{elhara2019coco}, which consists of 24 types of synthetic functions with customizable properties. For evaluation, we use the same \textit{test} set of functions used by \citet{chen2022towards} based on both datasets. \\
\textbf{Model and Training Details}: We use models provided by \citet{chen2022towards} which were pre-trained on the RealWorldData and BBOB datasets respectively. For specific details about the hyperparameters used to train these models and the model architecture, please see Appendix D.2 of \citet{chen2022towards}. In our experiments, we generate an initial seed set $C_{t}$ of 100 candidates that we sample from the OptFormer prior policy. We then use this set to obtain our rollout-generated horizon-aware candidate set, $C^{h}_{t}$. We fix $k = |C^{h}_{t}| = 50$ and experiment with rollout horizons $h = \{2, 3, 5, 10\}$.\\
\textbf{Baselines: } We compare our approaches to two baselines. 
\begin{enumerate}[leftmargin=*, topsep=0pt]
    \item \looseness=-1 Vizier: The Google Vizier Service \citep{golovin2017google} implements a GP-UCB (Gaussian process surrogate with upper confidence bound acquisition function) and is a baseline used in \citep{chen2022towards}. We choose this because it was the strongest performing algorithm amongst non-OptFormer models in \citet{chen2022towards}.
    \item OptFormer (EI): This is the base OptFormer policy augmented with the expected improvement acquisition function. This is the best policy found by \citet{chen2022towards} on the two datasets we consider.
\end{enumerate}

\section{Results and Discussion}
\label{section:results_and_disc}
In this section, we demonstrate that the changes that we have made on top of the OptFormer model lead to modest improvements in downstream performance.
\begin{figure}[t!]
    \centering
    \includegraphics[scale=0.2]{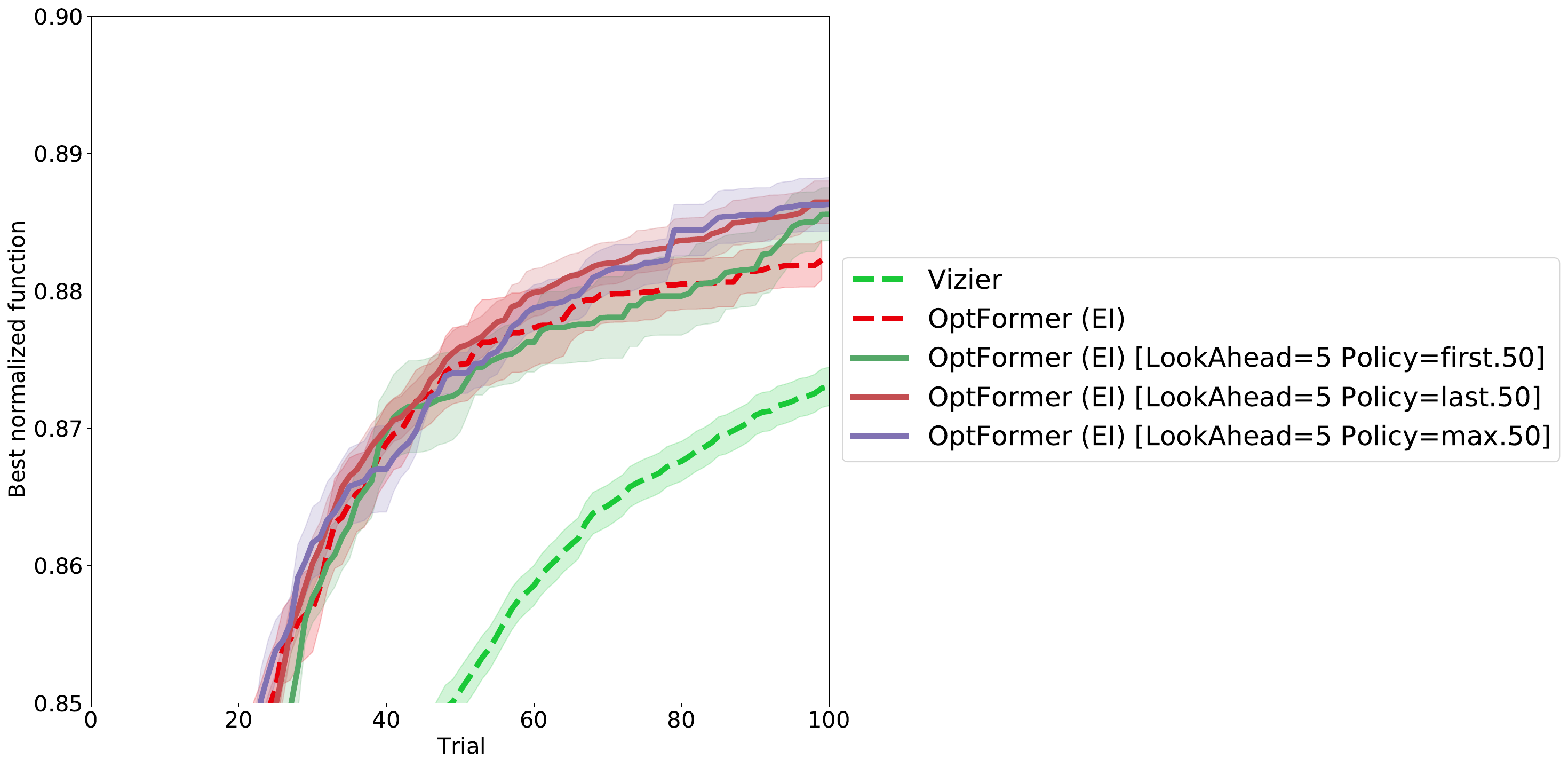}
    \caption{We evaluate augmenting OptFormer (EI) with our candidate generation strategies from Section \ref{subsect:cand_gen}. We use a lookahead horizon $h = 5$ which we find to be a reasonable length from ablations. We observe the following performance ranking towards the end of the trial : Max.50 $\approx$ Last.50 $>$ First.50 $\approx$ OptFormer (EI) $>$ Vizier}%
    \label{fig:vizier_candidate_generation}%
\end{figure}
\subsection{Rollout-aware candidate generation improves upon OptFormer (EI)}
Figure \ref{fig:vizier_candidate_generation} shows how refining $C_{t} \rightarrow C^{h}_{t}$ impacts the performance of our baseline model OptFormer (EI) on the RealWorldData dataset. As can be seen, we get modest improvements from using  Max.50 and Last.50. Our First.50 however mostly mimics the baseline OptFormer (EI) policy and does not show any gains until (arguably) the last few trials.

We posit that candidate generation leads to improvement in performance via the following mechanisms:
\begin{itemize}[leftmargin=*]
    \item Improved diversity of candidates: Members of $C_{t}$ and  $C^{h, F}_{t}$ are generated from the same distribution $P(\vx_{t} | m, \vh_{t - 1})$.  However for $C^{h, M}_{t}$ and $C^{h, L}_{t}$, we get higher diversity of samples since candidates are generated from different prompting sequences obtained from a diverse set of rollouts. More diverse candidates means we can explore more of the search space and thus increase our chances of finding a good hyperparameter configuration. Note that the First.k criterion produces a subset of $C_{t}$ and so does not increase the diversity of candidates as compared to the two other criteria. This could explain its failure to improve upon the baseline.
    \item Improved quality of candidates: As discussed in Section \ref{subsect:cand_gen}, taking the $\mathrm{argtopk}$ based on fantasized $\tilde{y}$ amongst rollout can be seen as a policy improvement step thus resulting in a better set of candidates. This improvement is only possible because OptFormer was trained to mimic a strong prior policy and its function approximation $P(y_t | m, \vh_{t - 1}, \vx_{t})$ is well calibrated \citep{chen2022towards}.
\end{itemize}

\begin{figure}[h!]%
    \centering
    \subfloat[\centering Full Plot]{{ \includegraphics[width=0.5\textwidth]{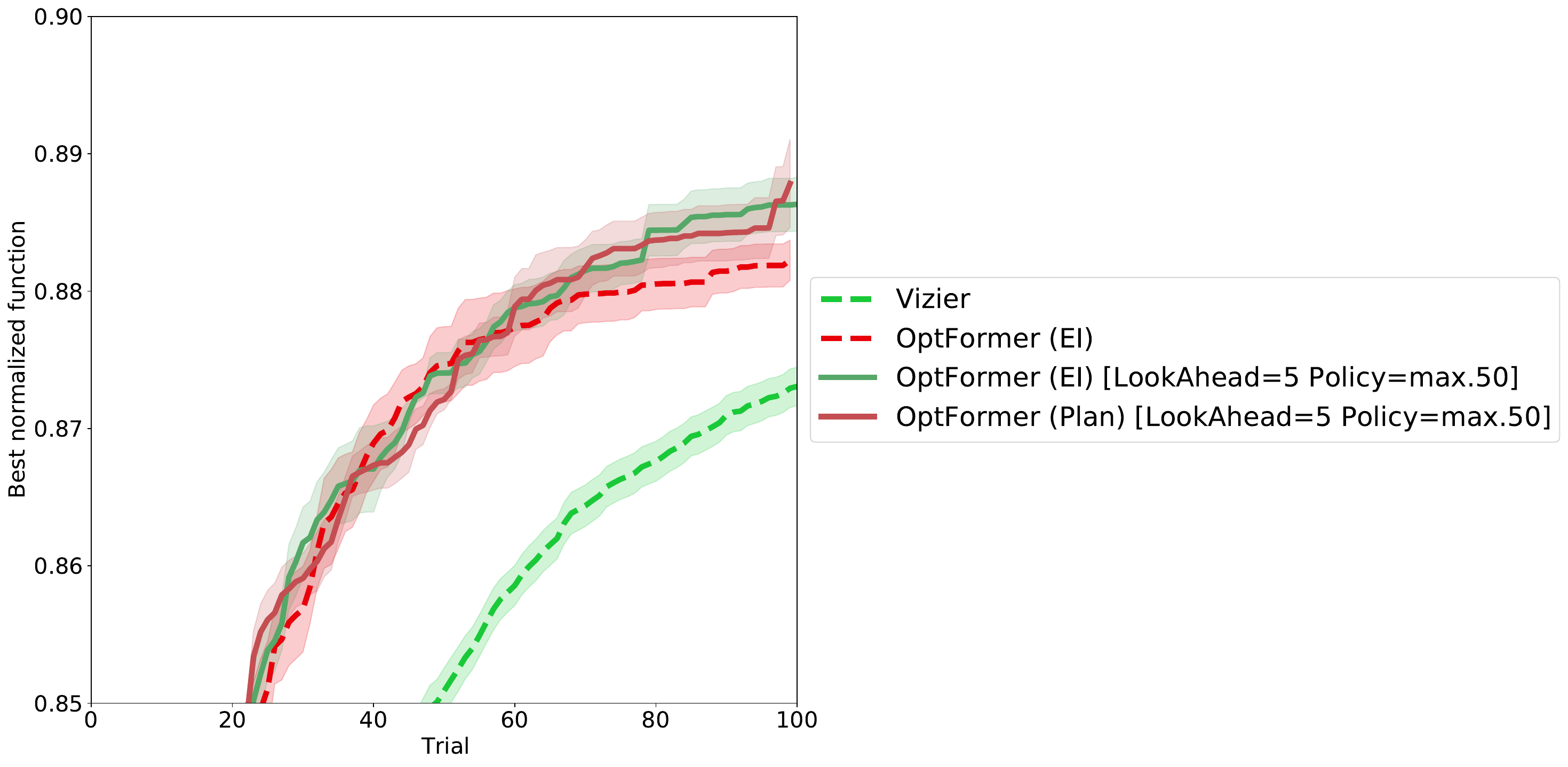} }}%
    \subfloat[\centering Zoomed In]{{ \includegraphics[width=0.5\textwidth]{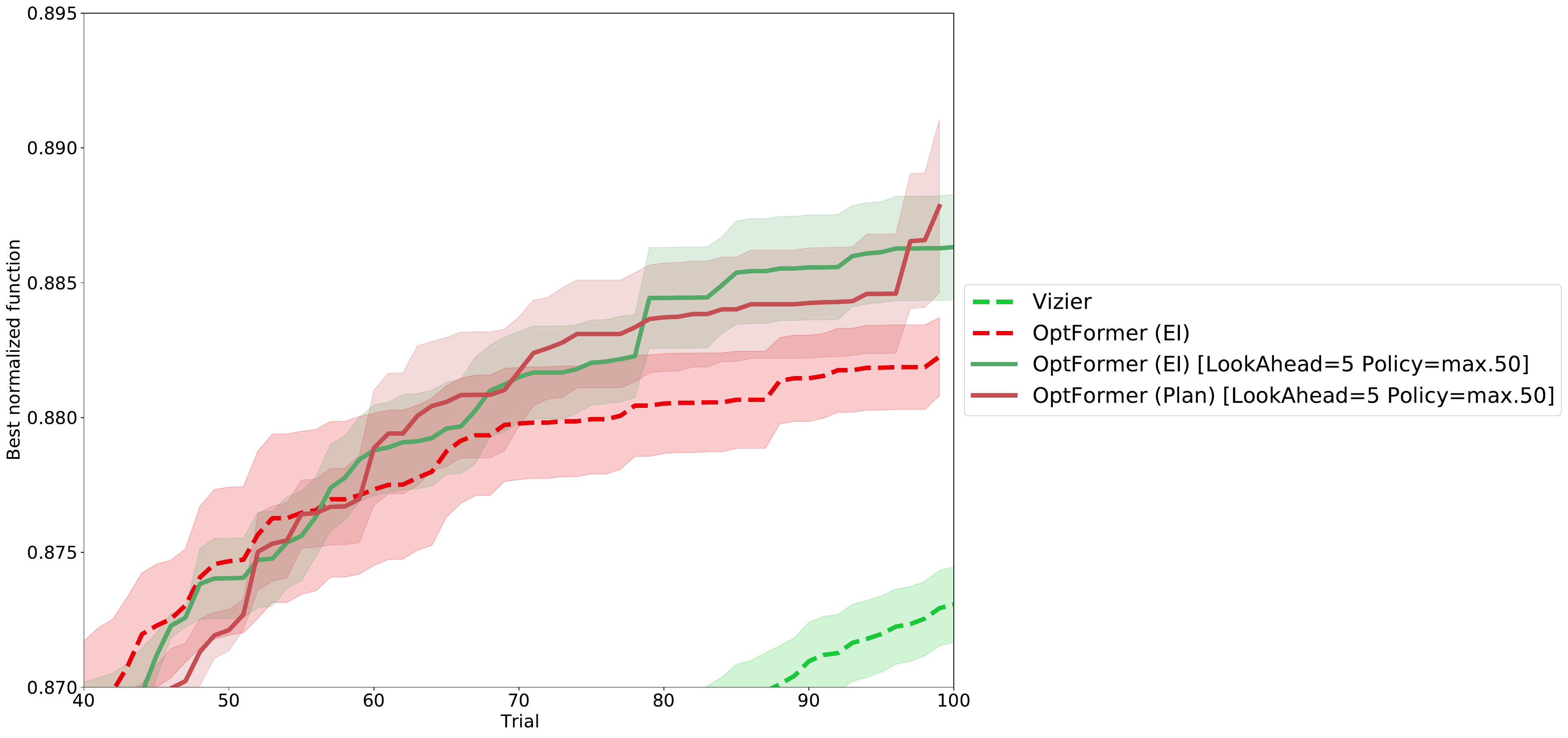} }}%
    \caption{We replace the EI acquisition function with our Planning-based acquisition function presented in \ref{subsect:cand_eval}. For each candidate, we generated $d = 64$ rollouts which we use to compute $\alpha_{\mathrm{Plan}}$. We do not observe any improvement from introducing this change.}%
    \label{fig:using_planning}
\end{figure}
\subsection{Planning-based acquisition function does not improve over rollout-aware candidate generation}
Encouraged by the gains from rollout-aware candidate generation, we introduce even more planning information during the candidate evaluation stage. We replace the expected improvement (EI) acquisition function ($\alpha_{\mathrm{EI}}$) with $\alpha_{\mathrm{Plan}}$ defined in Section \ref{subsect:cand_eval} but keep the Max.k candidate generation strategy. We compare these two approaches on the RealWorldData dataset in Figure \ref{fig:using_planning}.

From  Figure \ref{fig:using_planning}, we do not see significant gain in changing $\alpha_{\mathrm{EI}} \rightarrow \alpha_{\mathrm{Plan}}$. One hypothesis for why this might be the case is that the set $C^{h, M}_{t}$ already contains significant planning information thus rendering the new acquisition function irrelevant for taking into account future evaluations. We leave a more thorough investigation to future work.

\subsection{Long rollout horizon leads to compounding bias}
\begin{figure}[t!]
    \centering
    \includegraphics[scale=0.18]{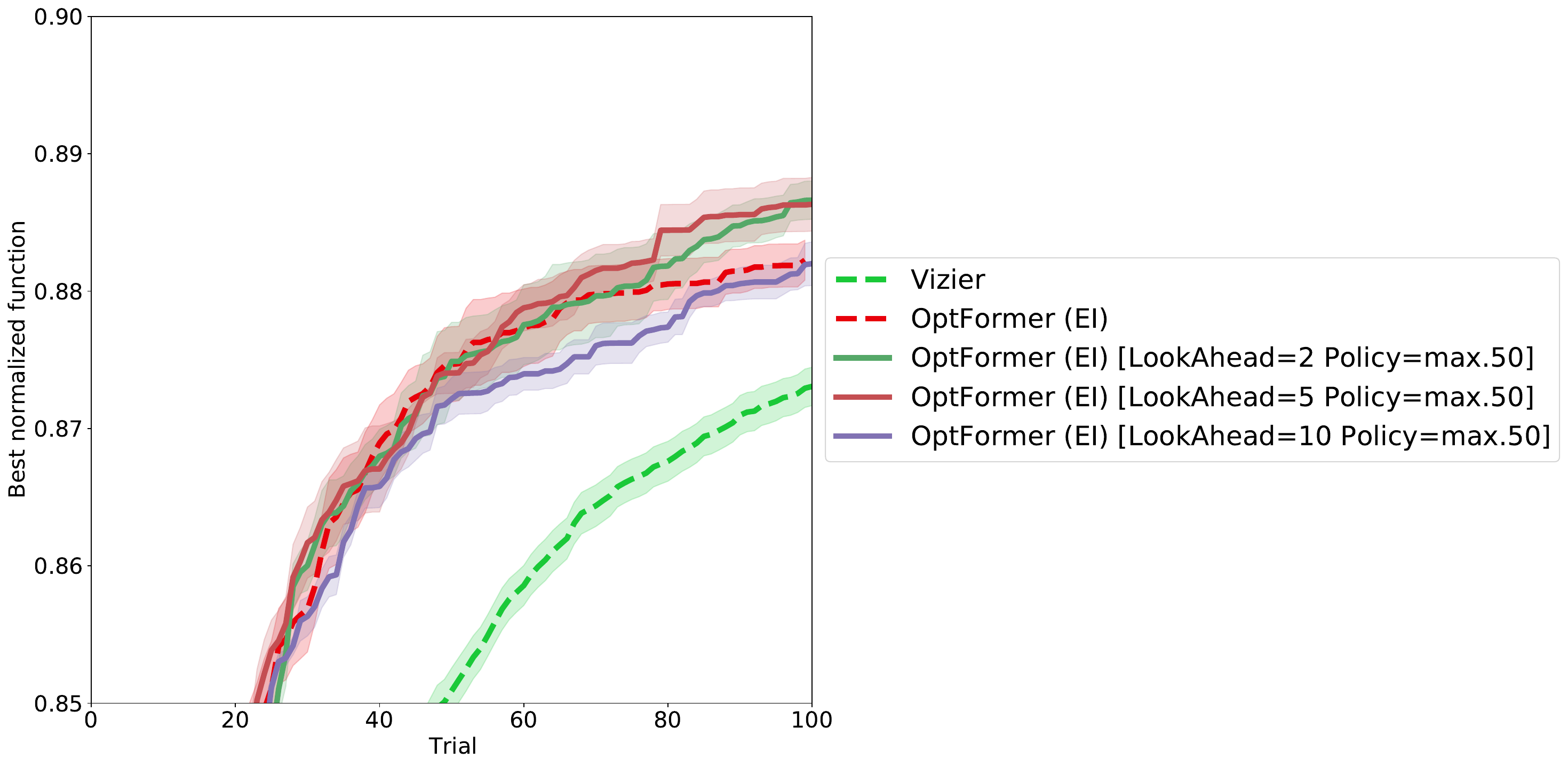}
    \caption{Ablation of $h$ on the RealWorldData dattset. Comparing look ahead horizons at k=50 for different Max.k based policies. In general, looking ahead too far into the future is not beneficial and actually degrades performance. A look ahead horizon of 2-5 seems to produce the best performance.}
    \label{fig:too_long_horizon}
\end{figure}

In order to understand the impact of the rollout horizon on performance, we fix our policy to [OptFormer (EI) - Max.k] but vary $h \in \{2, 5, 10\}$. We find in Figure \ref{fig:too_long_horizon} that using a very long horizon degrades performance. Note that whilst $P(\tilde{y}_t | m, \vh_{t - 1}, \tilde{\vx}_{t})$ is primarily based on an actual, observed trajectory $\vh_{t - 1}$, sampling at timestep $\ell$ of a rollout $P(\tilde{\vx}_{t + i} | m, \vh_{t - 1}, \tilde{\vh}^{\ell}_{t})$ requires a simulated trajectory $\tilde{\vh}^{\ell}_{t}$. The longer $\tilde{\vh}^{\ell}_{t}$ is, the more likely it deviates from the true $\vh^{\ell}_{t}$ thus leading to inaccurate decisions. Our ablation suggests a horizon of 2-5 works best with the OptFormer model. Note that previous approaches to HPO have to restrict $h \in \{1, 2\}$ in order to keep compute feasible. Our horizon is not limited by computation but by compounding bias. For future work, we could train OptFormer not just to imitate the base policy but also to predict rolled-out values more accurately so as to combat the issue of long horizon  bias.

\subsection{Performance on BBOB dataset}
We have demonstrated improvements to OptFormer on the RealWorldDataset. To validate the generality of our approach, we further test it on the BBOB dataset. Unfortunately we do not see much gain from our methods over OptFormer (EI) in this setting. As we conducted limited tuning of $h, k$ in this setting we believe more extensive tuning of these configurations could yield better results.

\section{Conclusion}
In this work, we have explored incorporating multistep planning into the recently proposed OptFormer HPO algorithm. We began by motivating why OptFormer is a suitable test-bed for exploring planning-based HPO and then presented different strategies to enable planning with OptFormer. Empirically, we show that our \textit{planning-horizon-aware} candidate generation strategies can lead to improvements when coupled with the expected improvement acquisition function. However, we note that our specialized acquisition function $\alpha_{\mathrm{Plan}}$ did not lead to desired levels of improvement over OptFormer. For future work, we hope to investigate the reason for this and use it to guide new proposals for planning-based acquisition functions.

\section{Acknowledgements}
We would like to thank Bobak Shahriari, Kazuya Kawakami, Caglar Gulcehre, Xingyou Song, Chansoo Lee, Zi Wang and Arnaud Doucet for helpful discussions at various points in producing this work. 
\newpage
{
\small
\bibliographystyle{unsrtnat}
\bibliography{references}
}

\newpage

\end{document}